\documentclass[letterpaper, 10 pt, conference]{ieeeconf}
\IEEEoverridecommandlockouts                              

\makeatletter
\let\NAT@parse\undefined
\makeatother
\usepackage[numbers,sort&compress]{natbib}

\usepackage{ctable}
\usepackage{amsmath,amssymb,amsfonts}
\usepackage{algorithmic}
\usepackage{graphicx}
\usepackage{textcomp}

\usepackage{lipsum}
\usepackage{tabularray}
\usepackage{multirow}
\usepackage{easytable}
\usepackage{pgf}
\usepackage{epsfig}
\usepackage[bookmarks=true]{hyperref}
\usepackage{subcaption}
\usepackage{cleveref}

\usepackage{soul,color}
\usepackage{verbatim} 

\usepackage{xcolor}

\def\BibTeX{{\rm B\kern-.05em{\sc i\kern-.025em b}\kern-.08em
    T\kern-.1667em\lower.7ex\hbox{E}\kern-.125emX}}

\usepackage[numbers]{natbib}

\usepackage{sparo_acronyms}
\usepackage{sparo_misc}

\title{\LARGE \bf
	Salience-guided Ground Factor for Robust Localization of \\ Delivery Robots in Complex Urban Environments
}

\author{Jooyong Park$^{1}$, Jungwoo Lee$^{2}$, Euncheol Choi$^{2}$ and Younggun Cho$^{2*}$
 	\thanks{This work was supported by the National Research Foundation of Korea(NRF) grant funded by the Korea government(MSIP) (RS-2023-00302589), Institute of Information \& communications Technology Planning \& Evaluation (IITP) grant funded by the Korea government(MSIT) (No.2022-0-00448) and Collabo R\&D between Industry, University, and Research Institute funded by Korea Ministry of SMEs and Startups in 2023 (00224459).} 
        \thanks{$^{1}$Jooyong Park is with the HL Klemove Corporation, Gyeonggi-do, and Dept. Electrical and Computer Engineering, Inha
University, South Korea.
            \texttt{jooyong.park@hlcompany.com} }
  	\thanks{$^{2}$Jungwoo Lee, $^{2}$Euncheol Choi and $^{2*}$Younggun Cho are with the Dept. Electrical and Computer Engineering, Inha University, Incheon, South Korea.
		\texttt{[pihsdneirf, cheol\textunderscore97]@inha.edu, yg.cho@inha.ac.kr } }%
}

\begin{document}

\renewcommand{\thetable}{\arabic{table}} 




\maketitle

\begin{abstract}
In urban environments for delivery robots, particularly in areas such as campuses and towns, many custom features defy standard road semantic categorizations. Addressing this challenge, our paper introduces a method leveraging \ac{SOD} to extract these unique features, employing them as pivotal factors for enhanced robot loop closure and localization. Traditional geometric feature-based localization is hampered by fluctuating illumination and appearance changes. Our preference for SOD over semantic segmentation sidesteps the intricacies of classifying a myriad of non-standardized urban features. To achieve consistent ground features, the \ac{MC-IPM} technique is implemented, capitalizing on motion for distortion compensation and subsequently selecting the most pertinent salient ground features through moment computations. For thorough evaluation, we validated the saliency detection and localization performances to the real urban scenarios. Project page: \href{https://sites.google.com/view/salient-ground-feature/home}{https://sites.google.com/view/salient-ground-feature/home}.
\end{abstract}

\section{Introduction} 

\begin{figure}[t] 
    \centering
    \includegraphics[width=\columnwidth]{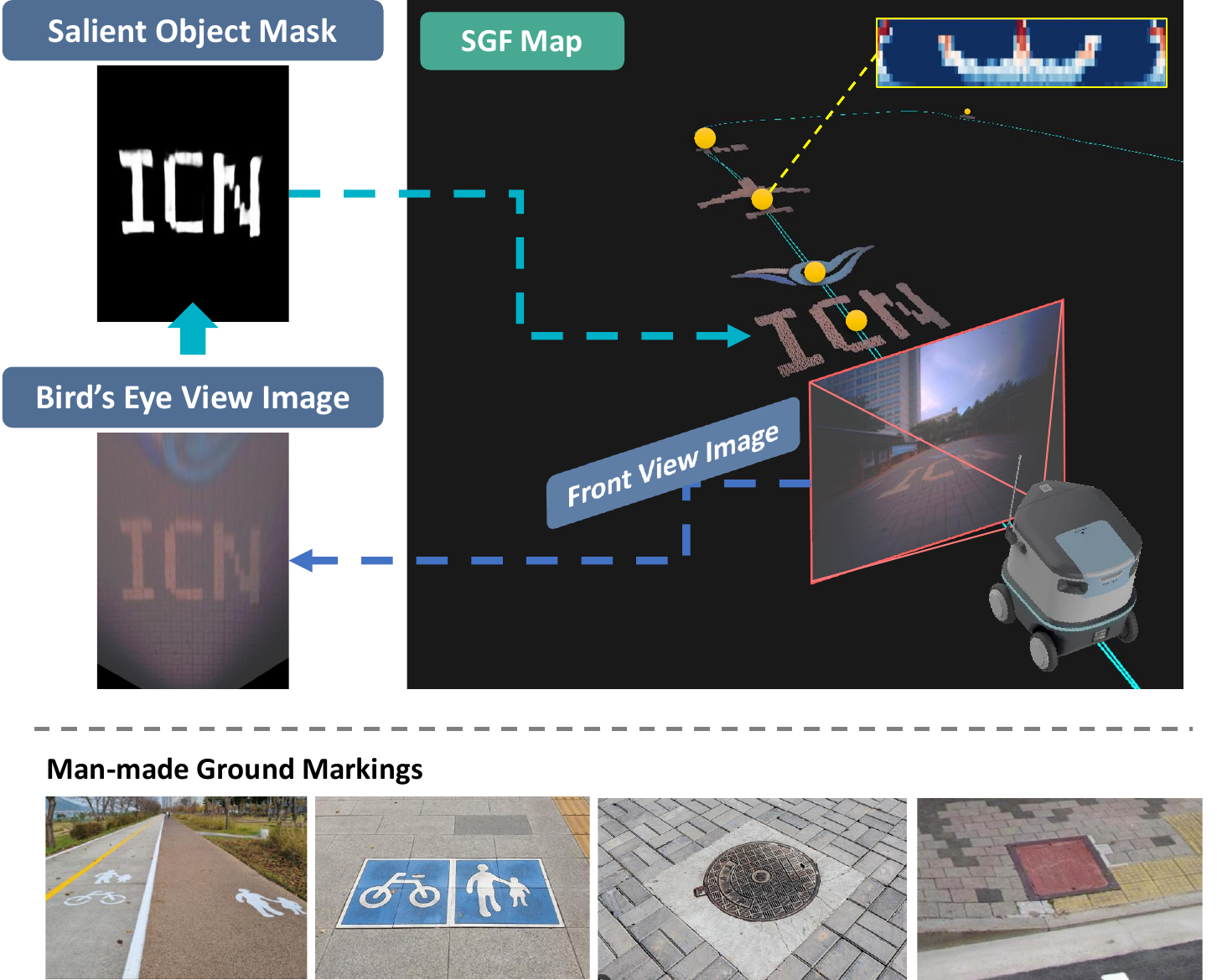}
    \caption{Illustration of the proposed method. We extract and describe the salient feature on the ground from the \ac{BEV}. It can be used as a loop factor to perform localization.} 
    \label{fig:main_figure}
\end{figure}

\begin{figure*}[t] 
    \centering
    \includegraphics[width=0.9\textwidth]{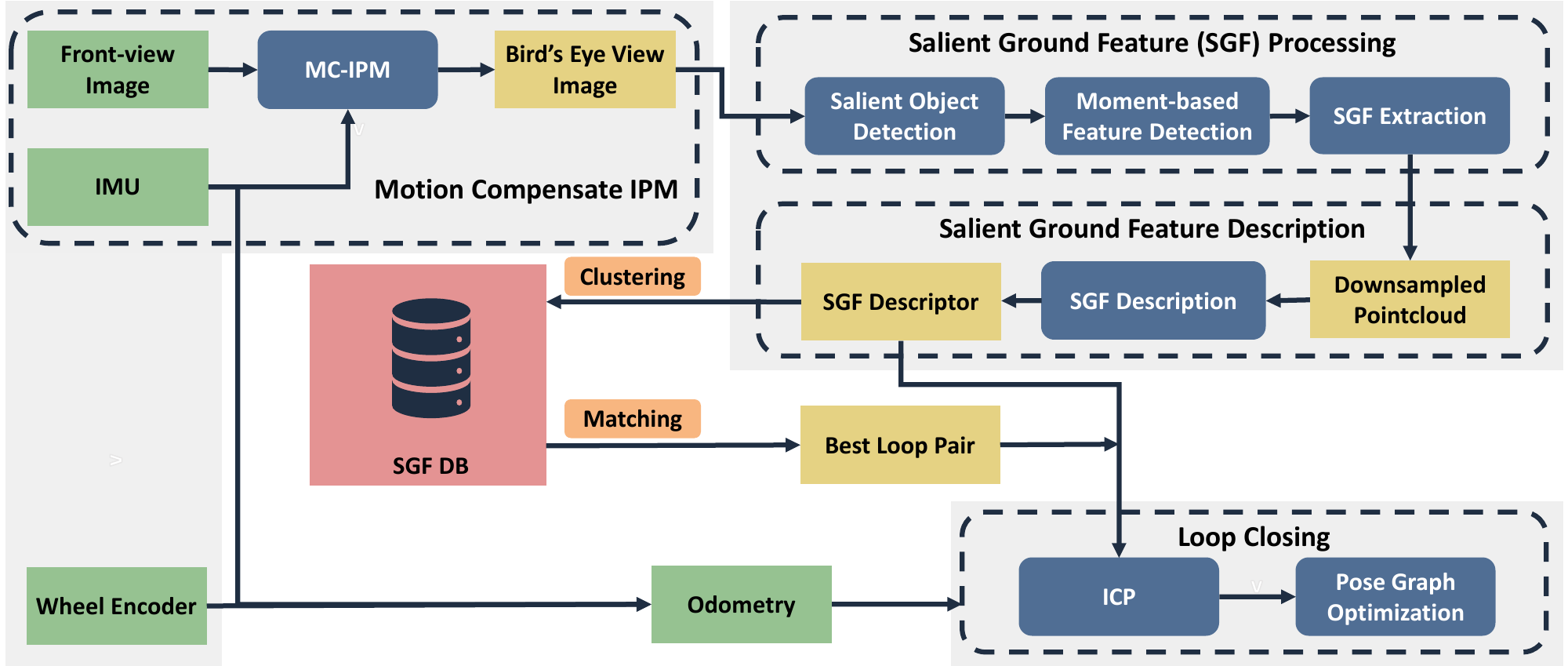}
    \caption{The overall pipeline of the proposed system. The diagram describes utilizing SGF for localization and loop closure.}
    \label{fig:overall_system}
\end{figure*}

In an era marked by the rapid integration of autonomous delivery robots into urban logistics systems, the robust and precise localization of these robots in complex urban environments has become an essential challenge. Accordingly, with advancements in Simultaneous Localization and Mapping (SLAM) \cite{campos2021orb, qin2018vins, zhang2014loam,shan2020lio} technology, last-mile delivery robots have been increasingly applied in various fields. For example, companies like Starship Technologies \cite{StarshipTechnologies}, Kiwibot \cite{Kiwibot}, Nuro \cite{Nuro}, and Neubility \cite{Neubility} provide delivery services based on environmental recognition through various sensors. 
 These delivery robots have different sensor configurations (such as RGB camera, LiDAR, GPS, etc.). Using various sensors allows the robot to be more aware of the environment, but they also increase the cost of the delivery robot. Therefore, vision-based SLAM methods received significant attention because of the preference for low-cost sensor configurations for commercial-level products. However, such methods suffer from illumination, perspective, and appearance changes in the long term. Furthermore, delivery routes in urban environments such as campuses are mostly weak-GPS or GPS-denied regions.

Interestingly, most man-made ground markings exhibit high visibility, making them particularly salient to human observers. Human localization systems often utilize these features as clues for topological localization. 
Motivated by this characteristic, we propose a localization system that can capture custom ground markings as the \ac{SGF} based on \ac{SOD}. \figref{fig:main_figure} describes the illustration of the proposed method. Even though other delivery robots may utilize a variety of sensor configurations, a front view single camera is commonly used. The proposed method first converts the monocular image to the \ac{BEV} with motion compensation. Then, we apply a \ac{SGF} detector and describe the \ac{SGF} in the descriptor vector. Finally, the delivery robot can correct the pose and localize on the map by utilizing \ac{SGF}. The details will be explained in \secref{sec:3}.


The proposed localization pipeline is as shown in \figref{fig:overall_system} and has the following contributions: 
\begin{itemize}
\item We propose a new salience-guided factor for loop closure and localization of delivery robots in urban scenarios. 
\item We present a robust \acf{IPM} to construct accurate ground features.
\item The proposed method incorporates salient feature detection to capture custom man-made features into the ground factors.
\item We validate our method in urban campus environments including various sequences, dynamic objects, and illumination changes. 
\end{itemize}



\section{Related Work}

In the field of autonomous robots, precise localization is one of the most critical tasks. 
As mentioned above, vision-based SLAM methods are cost-competitive but suffer from illumination, perspective, and appearance changes in the long term.
To address these challenges, various methodologies utilizing readily observable road markings such as lines and curb markers have been proposed \cite{schreiber2013laneloc, ranganathan2013light, lu2017monocular}. 
However, these approaches still have the same limitations as relying on traditional features.
Therefore, some approaches have been proposed using ground information as ground features \cite{jeong2017road, qin2021light}.  

{\bf{\ac{IPM}}}
The IPM algorithm, utilized for extracting robust ground features, is used in various applications \cite{tuohy2010distance, guo2014automatic, lin2012vision}. Traditional IPM techniques do not work well on sloped or rough surfaces \cite{ying2016robust}. To address this issue, some approaches \cite{zhang2014robust, nieto2007stabilization} use estimated vanishing points. In other works, \citeauthor{jeong2016adaptive} \cite{jeong2016adaptive} leverage camera pose information to obtain precise IPM images, introducing an extended IPM algorithm. More recently, \citeauthor{zhu2018generative} \cite{zhu2018generative} proposed a learning-based approach using synthetic datasets, while \citeauthor{bruls2019right} \cite{bruls2019right} presented a method based on real-world datasets, ensuring that IPM works well in real-world environments.

{\bf{Ground Feature-based SLAM}} Many works have leveraged ground features for robust localization in complex urban environments. Road-SLAM \cite{jeong2017road} exploits \ac{IPM} image and informative six classes of road marking that can be obtained by the machine learning approach. With the advancements in CNN-based image segmentation networks \cite{lee2017vpgnet, ahmad2017symbolic}, many approaches use this network to utilize ground features. 
The AVP-SLAM \cite{qin2020avp}, which leverages an image segmentation network \cite{ronneberger2015u} specifically trained for parking lots, enhances the accuracy of mapping and localization. \citeauthor{cheng2021road} \cite{cheng2021road} achieve efficient and accurate localization via compact semantic map with CNN-based sparse semantic visual feature extracting front-end. \citeauthor{zhou2022visual} \cite{zhou2022visual} proposed visual mapping and localization based on a compact instance-level semantic road marking parameterization.
However, these methods, relying on semantic image segmentation, suffer from pre-defined class limitation, which is unsuitable for complex and diverse environments. 

Delivery robots, operating in complex urban environments, face challenges such as unusual terrain (including paved roads, bumpy sidewalks, and obstacles like curbs) and dynamic occlusions from pedestrians and vehicles. They are also constrained by limitations in available sensors, including GNSS-denied scenarios and cheaper sensors. To address these problems, we introduce a \acf{MC-IPM} method in \secref{sec:3-mcipm}. We also demonstrate the extraction and utilization of undefined semantic ground features using a transformer-based \ac{SOD} model \cite{yun2022selfreformer} in \secref{sec:3-sgf-detection}, \secref{sec:3-sgf-description} and \secref{sec:3-loopclosing}.

\section{Method}
\label{sec:3}


\subsection{Motion Compensate Inverse Perspective Mapping}
\label{sec:3-mcipm}

We use the \ac{IPM} method to facilitate obtaining ground features from a monocular camera. Since \ac{IPM} assumes that the ground in front of the camera is flat, distortion compensation through local ground estimation is necessary in environments with uneven ground. To solve this problem, we introduce \ac{MC-IPM}, a post-processing process that measures the relative motion changes of the robot and compensates for them.

First, IPM compensation uses the difference in roll and pitch between poses. In the previous task, we improved the Adaptive \ac{IPM}\cite{jeong2016adaptive} into an Extended and Adaptive \ac{IPM} method that can compensate for both roll and pitch.
The pixel coordinates $(u, v)$ are converted into metric space $(\mathrm{c}, \mathrm{r})$ and projected to the compensated 3D point $(x_c, y_c, 0)$ of the camera center coordinates as follows,

\begin{equation}\label{eqn:roll_comp}
    \left[\begin{array}{c}
    \mathrm{c}^{\psi} \\
    \mathrm{r}^{\psi} 
    \end{array}\right]=
    \left[\begin{matrix}
        \cos(-\boldsymbol{\psi})  & -\sin(-\boldsymbol{\psi}) \\
        \sin(-\boldsymbol{\psi})  &  \quad\cos(-\boldsymbol{\psi}) 
    \end{matrix}\right]
    \left[\begin{array}{c}
    \mathrm{c} \\
    \mathrm{r} 
    \end{array}\right],
\end{equation}

\begin{align}\label{eqn:final_ipm}
    \begin{aligned}
        x_c &= h_c\cot(\alpha + \Phi(\mathrm{r}^{\psi}) + \boldsymbol{\theta}), \\ 
        y_c &= -x_c\bigg(\cfrac{\mathrm{c}^{\psi}}{f_m}\bigg)\cfrac{\cos(\Phi(\mathrm{r}^{\psi}) + 
        \boldsymbol{\theta})}{\cos(\alpha + \Phi(\mathrm{r}^{\psi}) + \boldsymbol{\theta})}.
    \end{aligned}
\end{align}

\begin{figure}[t] 
    \centering
    \includegraphics[width=\columnwidth]{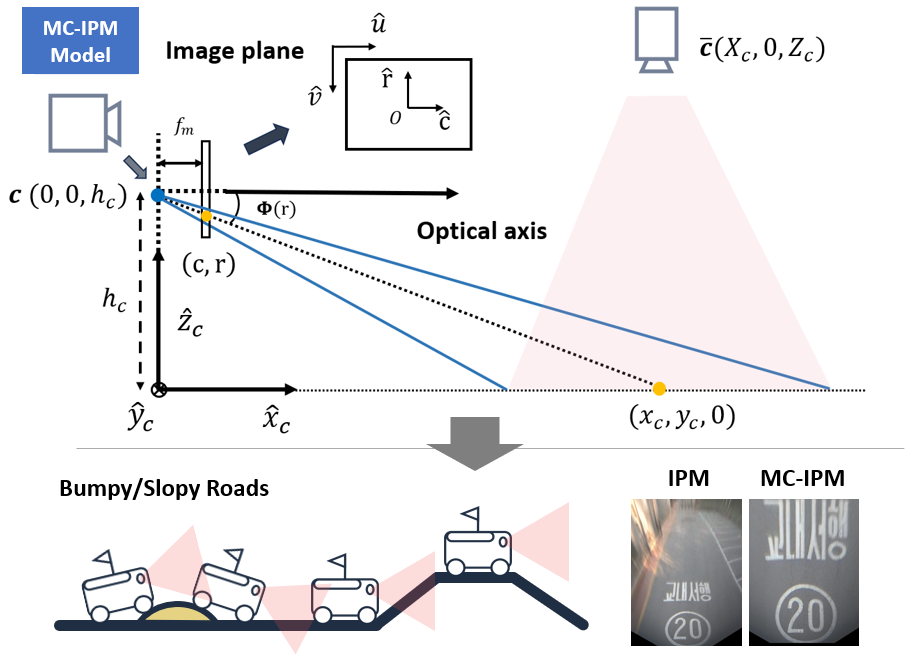}
    \caption{Illustration of an \ac{MC-IPM} model. A corrected \ac{BEV} image is generated from a compensated 3D point projected through the \ac{MC-IPM}. The lower right example is a compensation result of a situation crossing a speed bump.}
    \label{fig:IPM_figure}
\end{figure}

\noindent \equref{eqn:roll_comp} is the roll compensation process before projecting from the image plane to 3D space, and $\boldsymbol{\psi}$ is 
the magnitude of roll change. In \equref{eqn:final_ipm}, $h_c$ is the height of the front view camera ($c$), $f_m$ is the focal length in metric units, $\alpha$ is the angle between the optical axis and the ground, $\boldsymbol{\theta}$ is 
the magnitude of pitch and $\Phi(\mathrm{r}^{\psi})=\arctan(\mathrm{r}^{\psi}/f_m)$. The compensation formula for motion changes is detailed in \cite{jeong2016adaptive}. After that, we can obtain the compensated \ac{BEV} image by 2D projecting the 3D point to the virtual camera ($\bar{c}$) through the process as,


\begin{equation}
    \left[\begin{array}{c}
    x_{\bar{c}} \\
    y_{\bar{c}} \\
    1
    \end{array}\right]=    
    \left[\begin{matrix}
        0 & -1 & 0 \\
       -1 &  0 & X_c \\
        0 &  0 & Z_c
    \end{matrix}\right]
    \left[\begin{array}{c}
    x_c \\
    y_c \\
    1
    \end{array}\right],
\end{equation}

\begin{equation}
    \left[\begin{array}{c}
    u_{\bar{c}} \\
    v_{\bar{c}} \\
    1
    \end{array}\right]=\mathbf{K_{\bar{c}}}
    \left[\begin{array}{c}
    x_{\bar{c}} \\
    y_{\bar{c}} \\
    1
    \end{array}\right],
\end{equation}
 where  $(X_c,0,Z_c)$ and $\mathbf{K_{\bar{c}}}$ are position (camera center coordinates) and intrinsic parameters of virtual camera, respectively. The \ac{IPM} projection and the relationship between the front view camera and the virtual camera are shown in \figref{fig:IPM_figure}.

Compensation can be applied primarily in two cases: 1) temporary unevenness (e.g., speed bumps, cracks) and 2) uphill or downhill, as shown in \figref{fig:IPM_figure}.
First, store ${N}$-poses in a queue and calculate the average. 
Since pitch motion change mainly occurs in driving scenarios, the average calculation does not include poses greater than the pitch threshold. 
Subsequently, the relative difference between the average pose in the area and the robot's pose that needs to be compensated is passed as \ac{IPM} parameters $\boldsymbol{\psi}$ and $\boldsymbol{\theta}$. 
In this paper, we set the threshold for $N=50$ and relative pitch differences, $d_{\boldsymbol\theta}=0.025$ radians. The qualitative result of \ac{MC-IPM} compensation as shown in \figref{fig:IPM_figure}.



\subsection{Salient Ground Feature Detection}
\label{sec:3-sgf-detection}

In the general computer vision field, \ac{SOD} targets the segmentation of foreground objects in normal scenes. Motivated by this approach, we utilize general \ac{SOD} networks to capture representative ground features. For \ac{SGF} extraction, we adopt a SelfReformer\cite{yun2022selfreformer} network, a transformer-based model. However, the general SOD networks show difficulties on ground features. Therefore we re-design the training sets with a partial mixture of well-known road markings\cite{jang2021lane}.
To this end, we re-formulate SelfReformer into SelfReformer* that can capture the characteristics of \ac{SGF}. The input of the SelfReformer* is the \ac{BEV} image (3-channel) calibrated by MC-IPM and the \ac{SGF} mask (1-channel) is obtained as the output.

After obtaining the binary \ac{SGF} mask, we select the optimal \ac{SGF} based on the image moment that can capture the nature of the binary image. Rather than performing extraction for all query images, we perform an efficient way that extracts the most significant features. This also avoids the ambiguity of defining a feature as an SGF that has not fully appeared.

The moment-based \ac{SGF} detector uses a sliding window approach to find the optimal \ac{SGF} in 4-steps. 1) First, it determines the valid feature range, which is the range from when a feature appears to when it disappears. 2) Then, it calculates the Hu moment\cite{1057692} for all features in the valid feature range. Each element of the Hu moment consists of $\eta_{ij}$, which has robust properties in translation and scale, as follows,

\begin{equation}
    \mu _{{pq}}=\sum _{{u\in U}}\sum _{{v\in V}}(u-{\bar{u}})^{p}(v-{\bar{v}})^{q}I(u,v),
\end{equation}

\begin{equation}
    \eta _{{pq}}={\frac  {\mu _{{pq}}}{\mu _{{00}}^{{\left(1+{\frac {p+q}{2}}\right)}}}},
\end{equation}

\noindent where $I(\cdot,\cdot)$ represents intensity at pixel coordinates $(u, v)$, and $p$,$q$ is the order of moment. In addition, $\mu_{pq}$ is the central moment, which is invariant to translation, and $\eta_{pq}$ is obtained by normalizing the central moment through the zero-order central moment. The Hu moment $\mathcal{H}$ is calculated to be invariant to translation, scale and rotation as follows.


\begin{equation}
\mathcal{H}_i=\{h_i^1, h_i^2, \ldots, h_i^7\}, \quad i\in \mathcal{V}
\end{equation}

\noindent where $\mathcal{V}$ is all sets in a valid feature range, and the seven components of Hu moment follow \cite{1057692}. 3) Next, we calculate the distance between the current feature and the previous feature by shape matching using the Hu moments as follows, 


\begin{equation}
d_{hu}(\mathcal{H}_{i-1}, \mathcal{H}_{i})=\sum_{j=1}^7\left|h_{i-1}^j - h_i^j\right|, \quad i\in \mathcal{V}.
\end{equation}

\begin{figure}[t]
    \centering
    \includegraphics[width=0.48\textwidth]{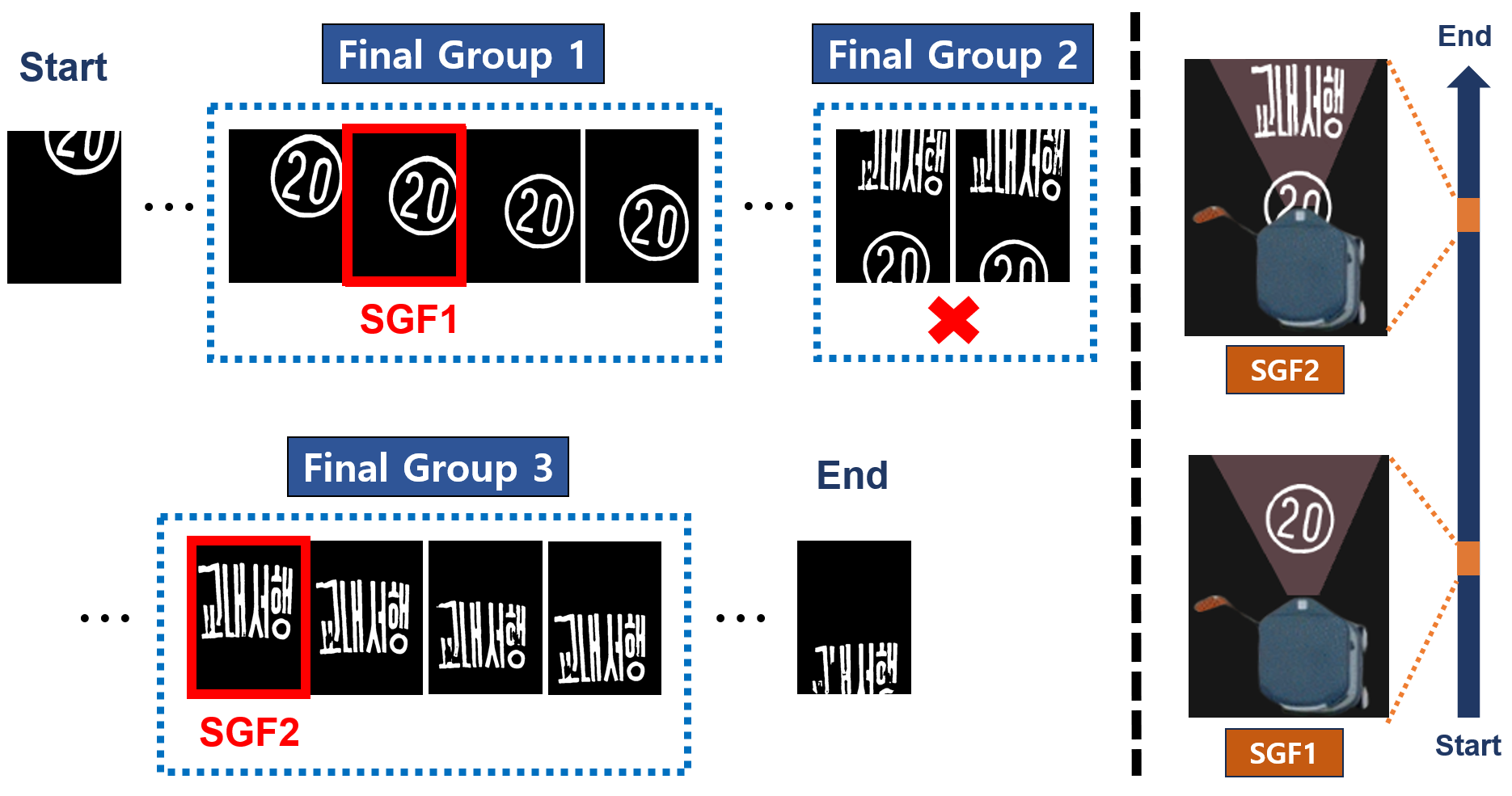}
    \caption{An example of a \ac{SGF} detected in a valid feature range. In the above case, two SGFs were detected in the three final groups, and not selected in final group 2 (boundary condition).}
    \label{fig:detector}
\end{figure}

\noindent Finally, 4) we select the  the group's most salient feature whose distance is closer than the threshold (=0.005). If the feature region is outside the four boundaries of the image, we do not select it. According to the three invariant properties of Hu moment, it can catch consecutive features that are highly correlated. An example is shown in \figref{fig:detector}.

\subsection{Salient Ground Feature Description}
\label{sec:3-sgf-description}






Once the optimal \ac{SGF} is selected, those regions are extracted into 3D space as follows,

\begin{equation}
    \left[\begin{array}{c}
        x_{sal} \\
        y_{sal} \\
        1
    \end{array}\right]=    
    \left[\begin{matrix}
        0 & -1 &  0 \\
       -1 &  0 & X_c \\
        0 &  0 & Z_c
    \end{matrix}\right]^{-1} \cdot \mathbf{K_{\bar{c}}}^{-1}
    \left[\begin{array}{c}
        u_{sal} \\
        v_{sal} \\
        1
    \end{array}\right],
\end{equation}

\begin{figure}[t]
    \centering
    \includegraphics[width=0.48\textwidth]{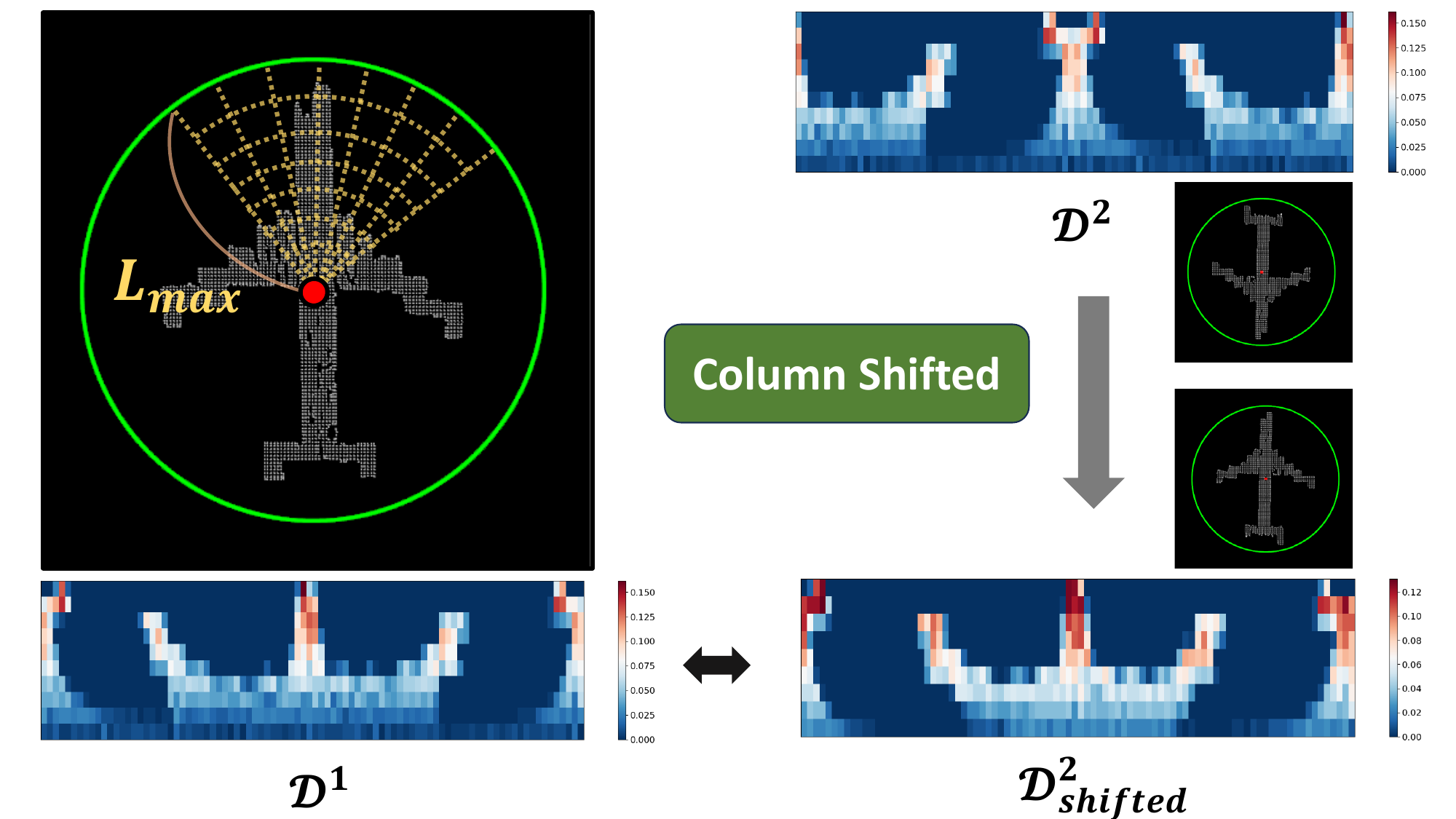}
    \caption{Illustration of \ac{SGF} description process and rotational invariance through column-shifted matching. The red dot is the center of the \ac{SGF} points, and the green circle is the radius $L_{max}$ boundary.}
    \label{fig:description}
\end{figure}

\noindent Afterward, downsampling is performed for \ac{SGF} points. To use the SGF as a loop factor, we create a descriptor for the \ac{SGF} points. We adopted the method proposed in scan context\cite{8593953}, which is a rotation robust description. Splits bins in azimuth and radial directions within the $L_{max}$ around the centroid of the \ac{SGF} points. In this paper, we used $L_{max}=2$, considering the average size of the ground feature. To perform a reverse loop closing, we set $N_s=90$, a central angle ($2\pi/N_s$), and $N_r=10$, a radial gap ($L_{\max }/N_r$). The description process is shown in \figref{fig:description}.

And then, \ac{SGF} descriptors are grouped into various features through online clustering. When the query \ac{SGF} descriptor $\mathcal{D}^q$ is determined, it is assigned to the closest group via cosine distance to the previous \ac{SGF} group's descriptor $\mathcal{D}^{sg}$ using column shifting as follows,

 \begin{equation}
     d(\mathcal{D}^q, \mathcal{D}^{sg}) = \frac{1}{N_c}\sum_{j=1}^{N_c} \left(1 - \frac{c_j^q \cdot c_j^{sg}}{\lVert c_j^q \rVert \lVert c_j^{sg} \rVert}  \right ),
\end{equation}

\noindent where $c_j^q$ and $c_j^{sg}$ are the j-th columns of $\mathcal{D}^q$ and $\mathcal{D}^{sg}$, respectively, and $N_c$ is the number of columns.
$\mathcal{D}^{sg}$ is then replaced by the mean of $\mathcal{D}$ in the group and is then used for comparison with the new $\mathcal{D}^q$. If the distance with all $\mathcal{D}^{sg}$ is greater than 0.7, it is assigned as a new \ac{SGF} group. 


\subsection{Loop Closing with SGF Factor}
\label{sec:3-loopclosing}



When the robot re-visits an area with the same \ac{SGF}, loop closing can be performed via \ac{SGF} association. Once \ac{SGF} is selected and the SGF group is determined, perform the \ac{ICP} matching with the closest descriptor within the group. In the reverse loop situation, we can provide the \ac{ICP} initials using the best column key obtained when computing the nearest descriptor. \ac{SGF} groups with symmetric properties (e.g., a circular manhole) are excluded from loop \ac{SGF} because they cannot provide an accurate pose as a result of \ac{ICP}. The global pose graph optimization with \ac{SGF} loop factor is as follows,

\begin{align}
    \begin{aligned}
         \mathcal{X}^*= \underset{\mathcal{X}}{\operatorname{arg}\min} \sum_{t} \lVert f(\mathbf{x}_t,\mathbf{x}_{t+1})-\mathbf{z}_{t,t+1} \rVert^2_{\Sigma_t} \\
         + \sum_{i,j\in \mathcal{L}} \lVert f(\mathbf{\mathcal{S}}_i,\mathcal{S}_{j})-\mathbf{z}_{i,j} \rVert^2_{\Sigma_{ij}},
    \end{aligned}
\end{align}

\noindent where $\mathbf{x_t}=\left[\mathbf{r}_t,\mathbf{t}_t \right]^T$ is the camera pose at $t$, $\mathcal{X}=\left[\mathbf{r}_0,\mathbf{t}_0, \ldots ,\mathbf{r}_t,\mathbf{t}_t\right]^T$ are all sequences, and $\mathbf{z}_{t,t+1}$ are relative pose between   camera frame at $t$ and $t+1$. $\mathcal{S}_i$ and $\mathcal{S}_j$ are \ac{SGF} pairs observed in different sequences and $\mathcal{L}$ is a set of all \ac{SGF} pairs. The function $f(\cdot,\cdot)$ estimates the relative pose.


\section{Experimental Results}

\subsection{Experimental Setups}

\begin{figure}[h!]
    \centering
    \includegraphics[width=0.82\columnwidth]{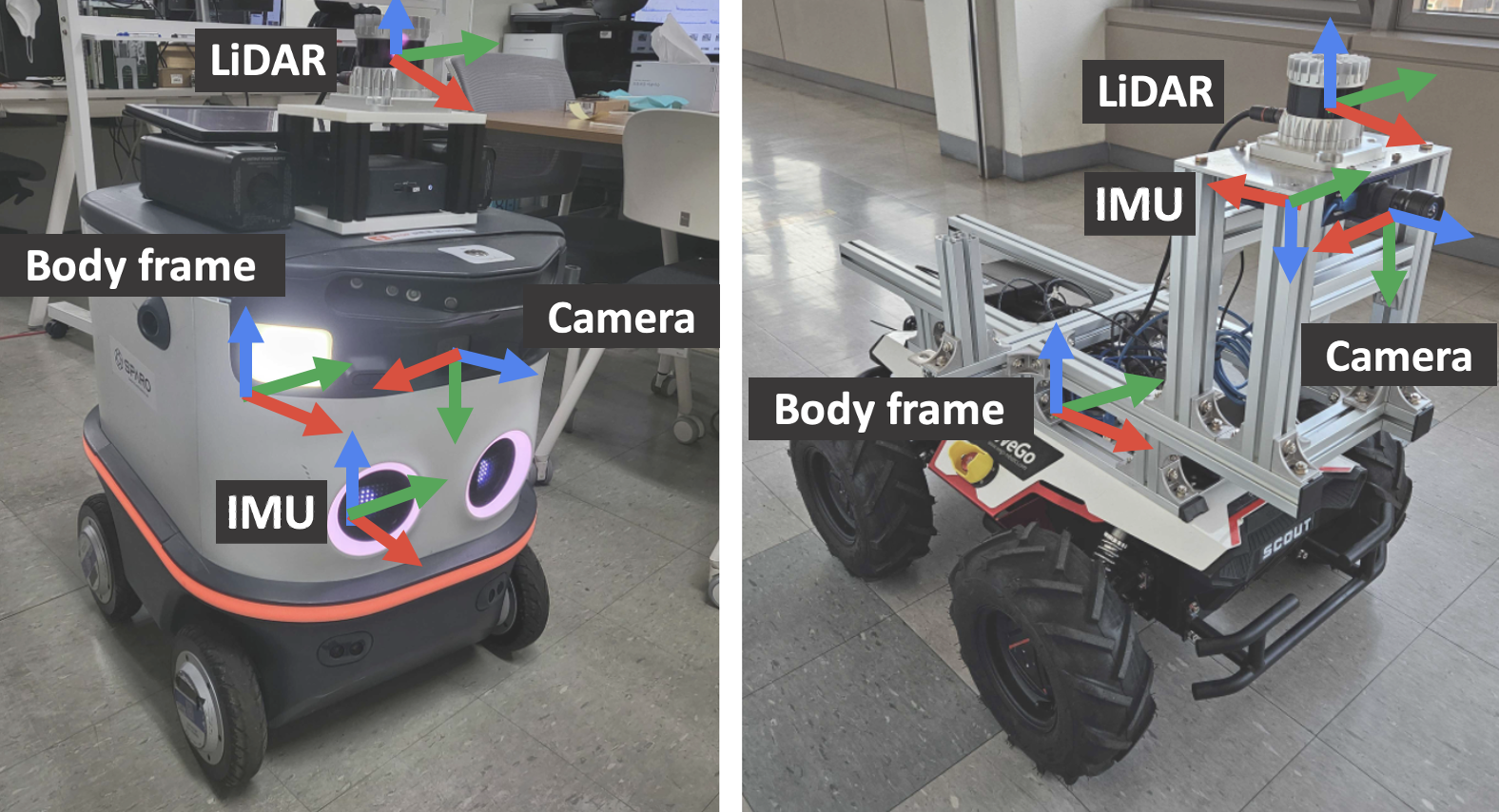}
    \caption{Robots were used in the experiments. (Left) A delivery Robot from Neubility \cite{Neubility}. (Right) A 4-wheeled Mobile Robot.}
    \label{fig:robot}
\end{figure}

In the experiments, we utilized two mobile robots: A real delivery Robot from Neubility \cite{Neubility} and a 4-wheeled mobile robot. \figref{fig:robot} represents the robots and sensors for evaluation. Both robots are equipped with wheel encoder, RGB camera, IMU and LiDAR. We constructed the ground-truth trajectories using well-known LiDAR SLAM methods \cite{xu2021fast, kim2022lt}. 

\begin{figure}[h!]
    \centering
    \includegraphics[width=0.9\columnwidth]{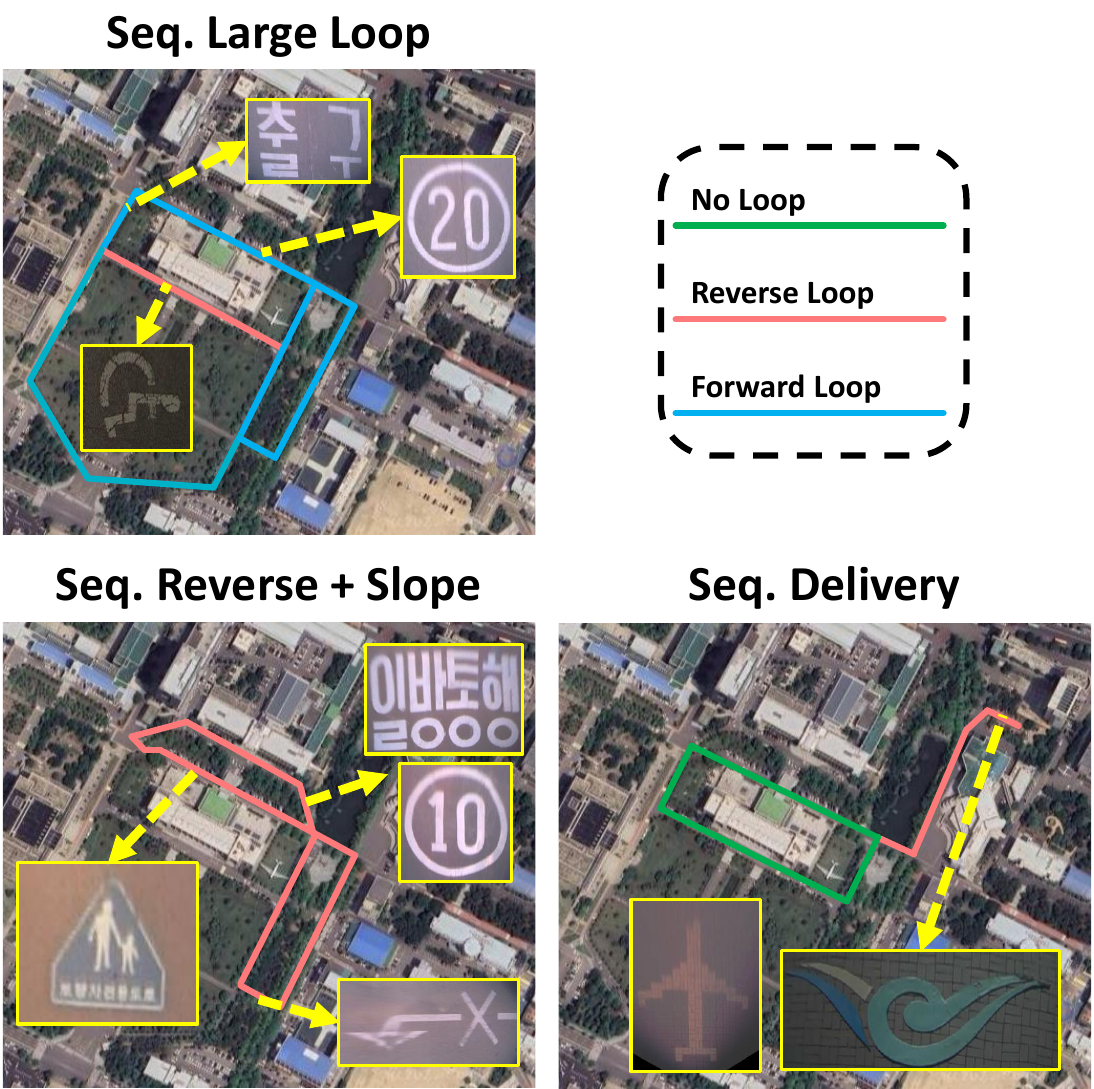}
    \caption{Test environment containing several representative SGFs. The legend indicates the type of the sequence.}
    \label{fig:seq_info}
\end{figure}

As described in \figref{fig:seq_info}, test sequences are composed of various driving scenarios on a campus because a campus environment includes many dynamic objects such as people and vehicles. Also, to evaluate the long-term operation, we evaluated the proposed method for the night sequences. \textit{Seq. Delivery} assumes a delivery route. In this scenario, the robot performs a delivery mission and then returns to its home. \textit{Seq. Reverse+Slope} is a scenario where all loops are reverse loops only. \textit{Seq. Large loop} covers about 2.7 kilometers, including both forward and reverse loops. 
For the concrete evaluation, we tested the proposed method in terms of saliency detection performance, feature detector score, and localization accuracy. 






\subsection{Salient Object Detection}

\begin{figure}[t!]
\centering
\vspace{-6.0em}
\includegraphics[width=1.0\columnwidth]{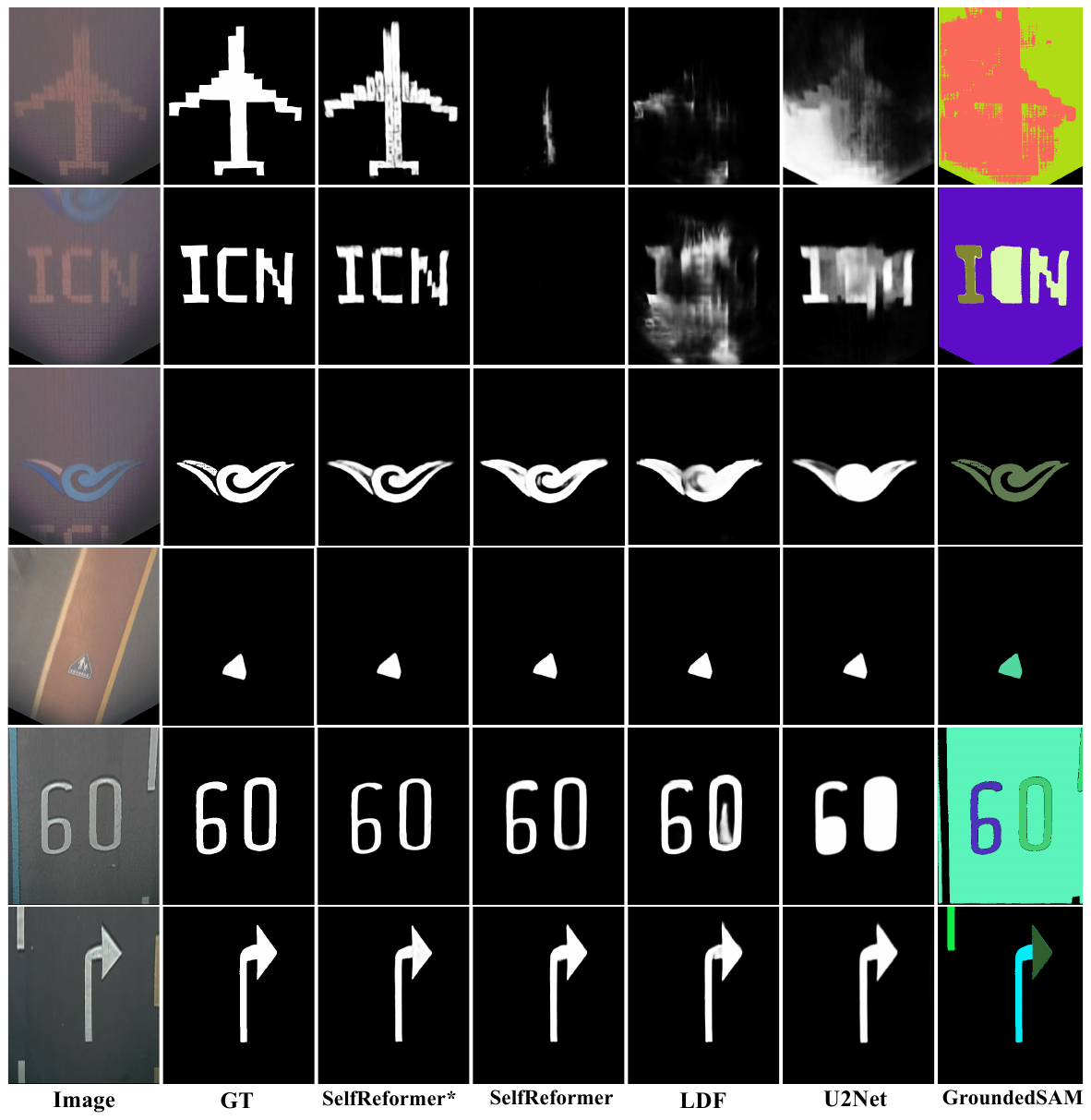}
\vspace{-7em}
\caption{\ac{SOD} results. Quantitative comparisons between SelfReformer*(Ours) and other SOD networks.}
\label{fig:sod_result}
\end{figure}

To test the performance of SelfReformer*, we compared the method to the well-known SOD networks \cite{wei2020label, qin2020u2} and the foundation model of semantic segmentation (GroundedSAM \cite{ren2024grounded}). 
In \figref{fig:sod_result}, SelfReformer* shows the best performance on saliency mask prediction. This result reports the validity of the training sets for man-made ground markings. Also, we found an interesting point on the result of GroundedSAM. Although GroundedSAM is known for great generality on visual perception, only some parts of the custom man-made markings can be represented by pre-defined semantics. On the other hand, SelfReformer* recognizes salient objects without pre-defined class information, so it can effectively recognize man-made features as well as standard markings in BEV images. For quantitative evaluation, we evaluated performance using well-known metrics MaxF ($F_m$), MAE ($M$), MeanF ($F_A$), S-measure ($S_m$) on the SOD task \cite{chen2018reverse, wei2020label, pang2020multi, wei2020f3net}. As shown in \tabref{tab:sod_result}, SelfReformer* outperforms all other models.

\begin{table}[h!]
\centering
\renewcommand{\arraystretch}{1.1} 
\resizebox{0.9\columnwidth}{!}{
    \begin{tabular}{l | c | c | c | c}
    \toprule
        {Methods}
        & {$F_m\uparrow$}   
        & {M\,$\downarrow$}
        & {$F_A\uparrow$}
        & {$S_m\uparrow$}
        \\
        \hline\hline
        
        
        SelfReformer*   & {\bf{.953}}  & {\bf{.032}}  & {\bf{.908}} & {\bf{.920}}    \\
        SelfReformer\cite{yun2022selfreformer}  & {.463}       & {.259}       & {.310}      & {.575}         \\
        LDF \cite{wei2020label}    & {.545}       & {.179}       & {.422}      & {.664}         \\
        U2Net \cite{qin2020u2}  & {.580}       & {.140}       & {.565}      & {.679}         \\

    \bottomrule
    \end{tabular}
}
\newline\textbf{}
\caption{Quantitative comparisons of SOD performance.}
\label{tab:sod_result}
\end{table}



\subsection{Localization Evaluation}
We present experimental results to validate the performance of the proposed auxiliary factor in the localization system. Since our algorithm is a vision-based localization method using both standard and man-made features in a complex urban environment, it is hard to compare directly with existing algorithms. Thus, we compared the results with Odometry and ORB-SLAM3. In ORB-SLAM3, we used a Monocular-IMU system to avoid scale ambiguity and manually perform scale correction on the outputs. We used evo \cite{grupp2017evo} to estimate \ac{ATE}.


\begin{figure}[t] 
    \centering
    \includegraphics[width=1\columnwidth]{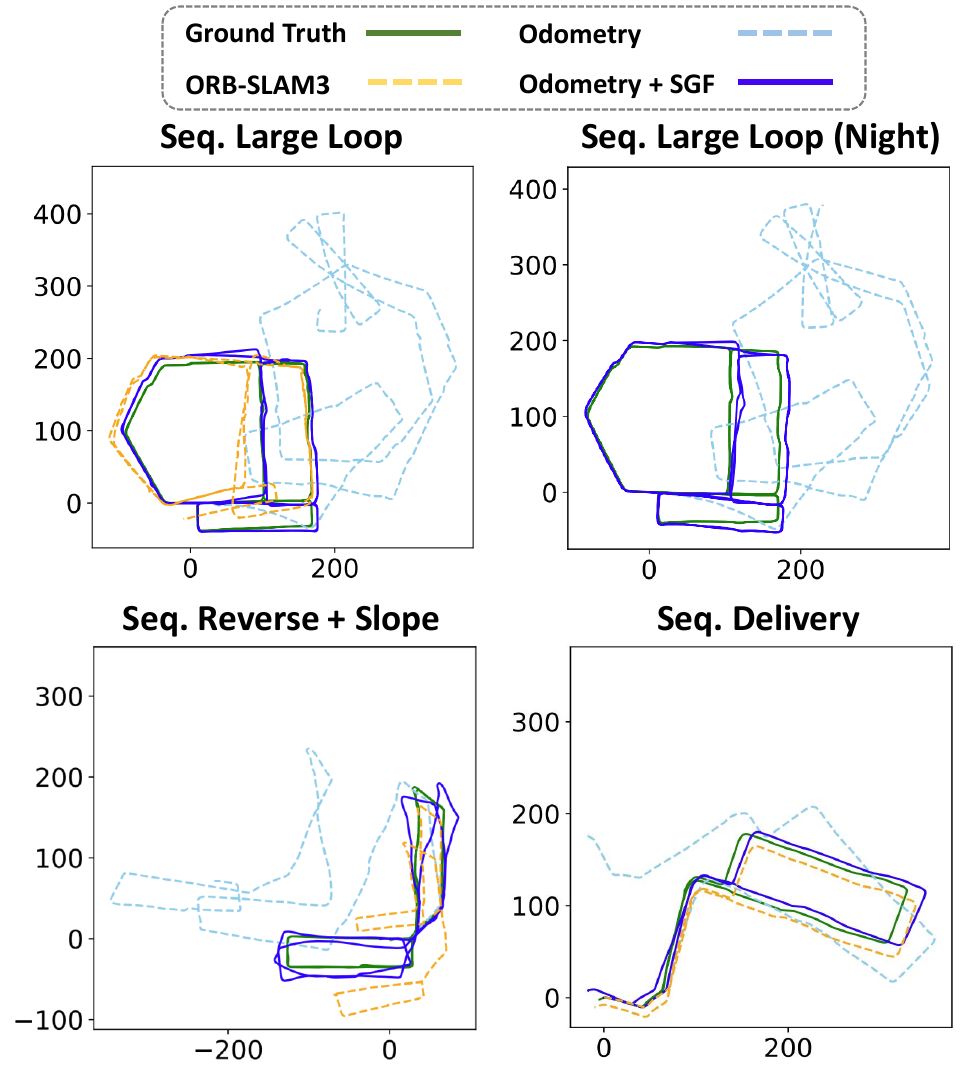}
    \vspace{-2mm}
    \caption{Results of adding \ac{SGF} factor with each sequence in campus scenario. Best viewed in color. }
    \label{fig:traj_plot}
\end{figure}

\begin{table}[h!]
\centering
\renewcommand{\arraystretch}{1.1} 
\resizebox{\columnwidth}{!}{
\begin{tabular}{l|r|r|r|r}
    \toprule
    \multirow{3}{*}{Methods}  &    \multicolumn{4}{c}{\bf{Sequence.}}
    \\ 
    \hline
     & \multicolumn{1}{c|}{Large Loop} 
     & \multirow{2}{*}{Reverse + Slope} 
     & \multirow{2}{*}{Delivery} 
     & \multicolumn{1}{c}{Large Loop }
    \\
     & \multicolumn{1}{c|}{(Day)}
     & {}
     & {}
     & \multicolumn{1}{c}{(Night)}
    \\
    \hline\hline
    Odometry    
        & {292.966}
        & {140.532}  
        & { 71.147}
        & {295.106}      \\
    ORB-SLAM3           
        & { 28.422}
        & { 48.546}  
        & { 19.828}
        & { - }   \\
    \bf{Odometry + SGF} 
        & { \textbf{9.482}}
        & { \textbf{14.384}}  
        & { \textbf{12.365}}
        & { \textbf{11.540}}  \\
    \bottomrule
\end{tabular}
}
\newline
\caption{Absolute Trajectory Error (ATE). (m)}
\label{tab:trajacc}
\end{table}

\figref{fig:traj_plot} shows how the proposed \ac{SGF} factor performs in campus scenario. The quantitative comparison can be found in \tabref{tab:trajacc}. 
For all sequences, We can see that the proposed method achieves a good performance compared to the traditional vision-based method, ORB-SLAM3. For \textit{Seq. Delivery}, both ORB-SLAM3 and the proposed method performed well, but we can see that our proposed method achieved lower errors with reverse loop detection. The \textit{Seq. Reverse + Slope} showed relatively low accuracy among all sequences because the SGFs were extracted from uneven ground. In the \textit{Seq. Large Loop}, ORB-SLAM3 lost tracking while driving. Comparing the trajectory before the tracking loss, we can see that the error was not corrected at the reverse loop point. 


\begin{table}[h!]
\centering
\renewcommand{\arraystretch}{1.1} 
\resizebox{\columnwidth}{!}{
    \begin{tabular}{l | c | c | c }
        \toprule
        {Sequences}& {Detection} & {Find loop pairs (Rev.)} & {Loop closed (Rev.)}
        \\
        \hline\hline
        
        Seq. Large Loop      & 79/94  &  33/34 (3/3)  &  31 (1)  \\
        Seq. Reverse+Slope    & 37/46  &  11/16 (11/16) &  7  (7)  \\
        Seq. Delivery        & 19/23  &    1/1 (1/1)  &  1  (1)  \\
          
    \bottomrule
    \end{tabular}
}
\newline\textbf{}
\caption{Quantitative results for \ac{SGF} detector, loop finding, and loop matching in each sequence.}
\label{tab:seq_metrics_result}
\end{table}

In \tabref{tab:seq_metrics_result}, each column quantitatively shows results for \ac{SGF} detector, loop finding, and loop matching by all sequences. The first column is the number of \ac{SGF}s found relative to the total number of \ac{SGF}s that should be detected in the path, the second column is the number of pairs grouped in the same \ac{SGF} group among all loop pairs and the last column is the number of \ac{SGF} loop matching was successful among the pairs obtained in the second column. Parentheses indicate the reverse loop cases. 
The proposed \ac{SGF} detector showed a detection success rate of nearly 80\% or higher, and the \ac{SGF} descriptor succeeded in loop matching in more than half, even in reverse loop cases.

We also tested the \ac{SGF} for the night-time sequences. Each robot is equipped with LED lights, and we applied Contrast Limited Adaptive Histogram Equalization (CLAHE) to enhance the contrast of dark images. As in \figref{fig:traj_plot}, the proposed method resulted in equivalent performance compared to the daytime. Note that ORB-SLAM 3 failed to optimize for both raw and enhanced night sequences. 


\subsection{Comparison of \ac{MC-IPM} and \ac{IPM}}
\begin{figure}[h!] 
    \centering
    \includegraphics[width=0.85\columnwidth]{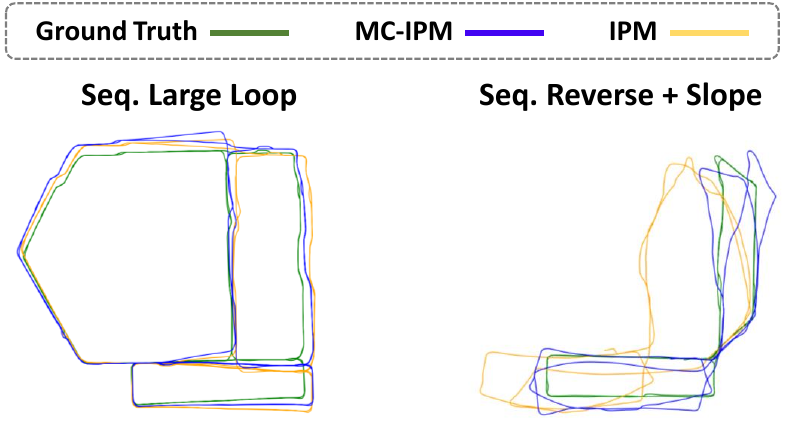}
    \caption{Qualitative comparison results of \ac{MC-IPM} and \ac{IPM}.}
    \label{fig:traj_mcipm}
\end{figure}

To evaluate the performance of MC-IPM, we compare with conventional \ac{IPM}. In uneven ground conditions, the \ac{IPM} cannot provide consistent information to the SGF detector, resulting in poor performance. For a more explicit comparison, we selected \textit{Seq. Large loop} and \textit{Seq. Reverse+Slope} with varying ground conditions and a large number of SGFs. The results are shown in \tabref{tab:MCIPM_IPM_result} and \figref{fig:traj_mcipm}.

\begin{table}[h!]
\centering
\renewcommand{\arraystretch}{1.1} 
\resizebox{0.9\columnwidth}{!}{
    \begin{tabular}{l | c | c }
    \toprule
        {Sequences}& {\ac{MC-IPM}} & {IPM}
        \\
        \hline\hline
        
        Seq. Large Loop      &  \textbf{9.482} $(m)$  &  12.567 $(m)$ \\
        Seq. Reverse+Slope   &  \textbf{14.384} $(m)$  &  67.625 $(m)$ \\
          
    \bottomrule
    \end{tabular}
}
\newline\textbf{}
.\caption{Comparison \ac{ATE} of \ac{MC-IPM} and \ac{IPM}.}
\label{tab:MCIPM_IPM_result}
\end{table}

\section{Conclusion}


In this paper, we propose a novel localization system that integrates standard and man-made features, one of the characteristics of complex urban structures, and utilizes them as SGF. To obtain more consistent \ac{SGF} in uneven ground conditions, we applied \ac{MC-IPM}, which utilizes the robot's motion. The \ac{SGF} factor was applied to a delivery robot, performing well in the campus environment and solving the reverse loop case. It also validated strong performance against changes in lighting and appearance throughout the day and night.
Our future work is to build a methodology that integrates \ac{SGF} into vision-based SLAM algorithms to perform better.

\bibliographystyle{IEEEtranN}
\bibliography{sparo_reference}

\end{document}